# Change Guiding Network: Incorporating Change Prior to Guide Change Detection in Remote Sensing Imagery[1]

Chengxi Han, *Student Member, IEEE*, Chen Wu, *Member, IEEE*, Haonan Guo, *Student Member, IEEE, Meiqi Hu, Student Member*, *IEEE, Jiepan Li, Student Member*, *IEEE*, Hongruixuan Chen, *Student Member, IEEE*

*Abstract*—The rapid advancement of automated artificial intelligence algorithms and remote sensing instruments has benefited change detection (CD) tasks. However, there is still a lot of space to study for precise detection, especially the edge integrity and internal holes phenomenon of change features. In order to solve these problems, we design the Change Guiding Network (CGNet), to tackle the insufficient expression problem of change features in the conventional U-Net structure adopted in previous methods, which causes inaccurate edge detection and internal holes. Change maps from deep features with rich semantic information are generated and used as prior information to guide multi-scale feature fusion, which can improve the expression ability of change features. Meanwhile, we propose a self-attention module named Change Guide Module (CGM), which can effectively capture the long-distance dependency among pixels and effectively overcomes the problem of the insufficient receptive field of traditional convolutional neural networks. On four major CD datasets, we verify the usefulness and efficiency of the CGNet, and a large number of experiments and ablation studies demonstrate the effectiveness of CGNet. We're going to open-source our code at https://github.com/ChengxiHAN/CGNet-CD.

## I. INTRODUCTION

THE technique of finding variations in the condition of a thing or phenomenon by watching it at different periods is change detection (CD) [1]. Very high-resolution (VHR) remote sensing images have been developed, making the change detection tasks accessible to a wide range of urban extension studies [2], land uses land cover analysis [3], environmental monitoring [4], and disaster assessment applications.

Conventional methods typically detect changes based on single-pixel changes, for instance, the image difference method [5], image ratio method [6], etc. Furthermore, there are also some methods based on feature transformation, such as change vector analysis [7], multivariate change detection [8], principal component analysis [9], slow feature analysis (SFA) [10] etc. For the above methods, it is crucial to define an appropriate decision function to distinguish whether the region has changed or not. Such classification-based [11] methods have been widely employed as an alternative to threshold-based methods in recent decades with the emergence of machine learning. Deep learning, a subset of machine learning, is widely used in computer vision and natural language processing. There have been fruitful research results in areas such as speech recognition. Computer vision, a popular deep learning research area, has produced a significant amount of work on remote sensing imagery classification [12], semantic segmentation [13], [14], object detection [15] and other tasks [16] [17]. As a result, one of the hottest areas of study in the world of remote

[1] This work was supported in part by the National Key Research and Development Program of China under Grant 2022YFB3903300 and 2022YFB3903405, and in part by the National Natural Science Foundation of China under Grant T2122014, 61971317, 62225113 and 42230108. (*Corresponding author: Chen Wu.*)

Chengxi Han, Chen Wu, Haonan Guo, Meiqi Hu and Jiepan Li are with the State Key Laboratory of Information Engineering in Surveying, Mapping and Remote Sensing, Wuhan University, Wuhan 430079, China (e-mail: chengxihan@whu.edu.cn; chen.wu@whu.edu.cn; guohnwhu@163.com; meiqi.hu@whu.edu.cn; jiepanli@whu.edu.cn).

Hongruixuan Chen is with the Graduate School of Frontier Sciences, The University of Tokyo, Chiba 277-8561, Japan (e-mail: Qschrx@gmail.com).

sensing is the application of deep learning algorithms from the field of computer vision to remote sensing imagery change detection[18], [19].

Convolutional Neural Networks (CNN) are the cornerstone of deep learning in computer vision. In order to extract image features, the convolutional neural network is quite effective. To improve image recognition, deep CNN can extract high-order semantic characteristics from images. remote sensing image change detection is the same as ground object classification, which necessitates pixel-level classification. To the best of our knowledge, we discover that the two primary types of deep learning change detection models are transformer-based and CNN-based (including attention mechanisms). All of these techniques rely on CNN's robust information extraction capabilities.

First, we introduce the main methods of using CNN as the base model. FC-EF [20], FC-Siam-conc [20], and FC-Siam-diff [20] were proposed by Daudt et al. at the same time. FC-EF [20], the abbreviation of fully convolution early fusion, is established on the basis of traditional U-Net. Fully convolution early fusion, or FC-EF [20], is based on conventional U-Net and is referred to as such. Early Fusion is abbreviated as EF. Before sending two input images to the network, the EF structure concatenates them together. They may be thought of as various picture channels. Feature maps from the two encoder branches and the corresponding layer of the decoder are connected by the FC-Siam-conc (Fully Convolutional Siamese-Concatenation) network [20]. In comparison, FC-Siam-diff (Fully convolutional Siamese-Difference) [20] makes a skip connection with the relevant layer of the decoder after first determining the absolute value of the difference between the feature maps of the two decoder branches. In order to create a network model, Mou et al. [21] combined Siamese CNN and LSTM to extract the spatial and spectral features of the image through CNN and then used LSTM to extract the temporal features. As a result, it is able to outperform some mainstream deep learning methods in terms of change detection. Additionally, some studies enhance the detection accuracy of the model from two perspectives of training strategy and network structure by including attention mechanisms and a deep supervision approach to the fully convolutional network. In order to address the issues of sample imbalance and low model accuracy in change detection, Han et al. proposed a HANet [22], which contains several attention mechanisms named HAN and numerous sampling techniques named PFBS. Using highly representative deep features in a deeply supervised difference discrimination network for change detection, Zhang et al. proposed a deeply supervised image fusion network (IFNet) [23]. In a Siamese-based spatial-temporal attention neural network (STANet), Chen et al. constructed a CD self-attention mechanism to express the spatial-temporal interactions [24]. Fang et al. designed SNUNet-CD [25], which is a Siamese network with several connections and is combined with NestedUNet for CD. A new deep multiscale Siamese network (MSPSNet) [26] including self-attention and concurrent convolutional structure was designed by Guo et al. Some researchers used this technique in the area of change detection because of UPerNet's superior feature extraction capability [27]. However, these methods still need to be improved when extracting the feature details of change information.

Subsequently, some scholars have also proved that the transformer-based method performs better in the field of CD. A deep learning model called Transformer [28] is fully based on the self-attention process because of its applicability to parallel computing and the complexity of the model itself. It can directly lead to its accuracy and performance being higher than the previous popular CNN-based methods. After the transformer has become popular in natural language processing, many scholars tried to introduce it into computer vision tasks. ViT(vision transformer) [29] is a model proposed in 2020 that directly applies Transformers to image classification. Its best model achieves 88.55% accuracy on ImageNet1K, which shows that Transformer is indeed effective in the CV domain, and the results are quite impressive. The Swin Transformer network is another collision of Transformers proposed in 2021 in the field of vision. Researchers in remote sensing have also introduced transformers into change detection tasks, such as SwinSUNet [30], BIT [31], Change Former [32], RSP-BIT [33], TransUNetCD [34], etc. Swin transformer blocks are used as the building blocks for the encoder, fusion, and decoder found in SwinSUNet [30]. A bitemporal image transformer called BIT [31] is used to effectively and efficiently effectively and efficiently model settings in the spatial-temporal domain.

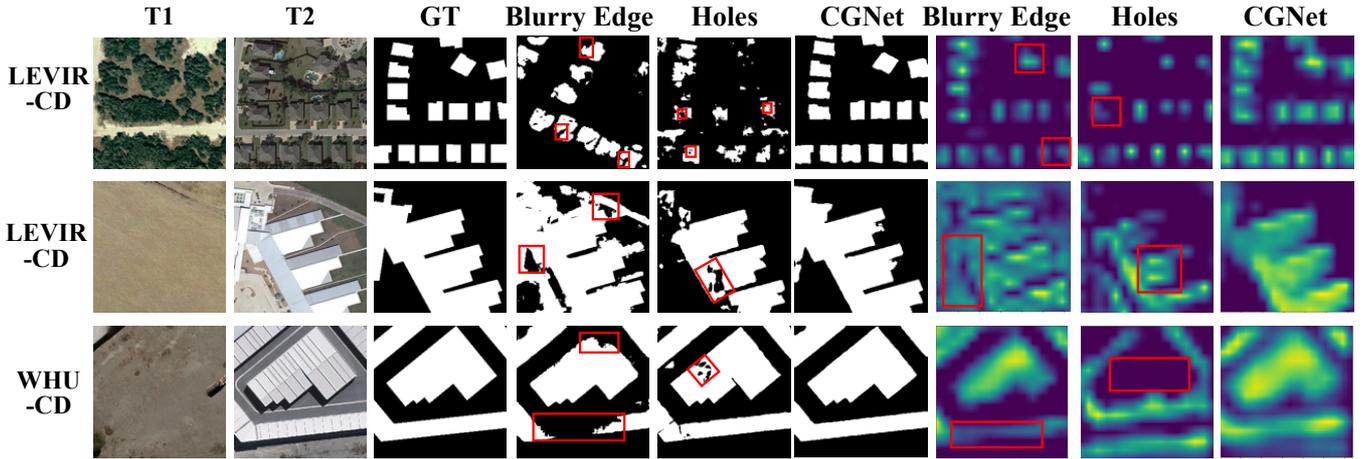

**Fig. 1.** Incompleteness of some edge pixels and the phenomenon of holes inside the changed areas.

To efficiently express the multi-scale and long-range features necessary for the precise CD, Change Former[32] combines a hierarchically structured transformer encoder with Multi-Layer Perception (MLP) decoder in a Siamese network architecture. RSP-BIT [33] introduces remote sensing pretraining (RSP) into BIT [31]. An end-to-end encoding-decoding hybrid transformer approach for CD called TransUNetCD [34] combines the benefits of transformers with UNet. However, some Transformer-based methods still have room for improvement in training speed while ensuring accuracy, which limits their practical application. Although the performance of transformer-based methods is better to some extent, these models are very large in the number of parameters and consume a lot of computing resources to train.

We've visualized the results, as seen in Fig. 1, to better understand how these models performed. We can see that in the previous methods, both the feature map and the binary map have problems such as edge breakage and internal holes. This is mainly caused by the insufficient context extraction ability of the model, the small receptive field, and the insufficient multi-scale fusion. There is no doubt that if we can use more complete deep features to guide the extraction of change features, we can make the effect of model extraction better.

Therefore, we propose a change guiding network (CGNet), which employs a deep change guiding feature as previous knowledge to direct multi-scale feature fusion, in order to more effectively handle these issues. And we propose a CGM module to get a big receptive field which is a self-attention module. The inaccurate edge detection and internal holes problems can be solved well by CGNet with a CGM module. The following is a summary of this work's key contributions:

1) To address the issue of inadequate expression of change characteristics in the conventional U-Net structure used in previous approaches, which results in inaccurate edge detection and internal holes, we propose a change guiding network, CGNet.
2) To enhance the expression capability of change features, we generate change maps from deep features with rich semantic information and use them as prior information to guide multi-scale feature fusion. Moreover, the proposed CGM module is a self-attention module, which can effectively compensate for the long-distance dependency among pixels and effectively overcome the problem of the insufficient receptive field of traditional convolutions neural networks.
3) On four CD datasets, extensive experimentation and ablation studies have been performed to confirm the usefulness and practicability of our approach. The program will be released as open source in an effort to further the remote sensing CD study.

The remainder of the article is organized as follows. The specifics of our proposed CGNet approach are introduced in Section II. The experimental setup and analysis of the results in detail and the ablation study are presented in Section III. Last, the paper's conclusion will be drawn in Section IV.

## II. CGNET

We will introduce the general CGNet architecture, the CGM module, and the model specifics in this section.

### A. Overall architecture

The U-Net structure is a classical and effective structure in extracting change information, but it still suffering from the problem of insufficient edge information extraction and internal holes. Therefore, we improve the conventional U-net architecture and propose a hierarchical feature network CGNet. Fig. 2 presents a schematic of the overall framework of the proposed CGNet. In order to extract the changing features of the bi-temporal image from coarse to fine, we first employ the VGG-16 network as the foundation in the encoder section. We use five VGG-16 blocks with batch normalization to represent the operation of VGG16_BN from 0-5,5-12,12-22, 22-32, and 32-42 layers, respectively. Then in the Decoder part, we refine the features by concatenating the features at different depths together and then passing the result through the designed convolutional block. The integration of multi-scale features is the most significant component of our network, which mainly generates the change map from the deep features with rich semantic information and uses it as prior information to guide self-attention to improve the expression ability of change features. Multi-scale feature fusion is guided by prior information. We propose a self-attention module CGM module, which can effectively compensate for long-distance dependency among pixels and effectively overcome the problem of the insufficient receptive field of traditional convolutional neural networks. We put three CGMS in different feature extraction stages of the network, namely CGM 4, CGM 3, and CGM 2. In fact, the number of 1, 2, 3, and 4 in the network represent different stages in the feature extraction process. The numbers of CGM and Convolutional Blocks are set according to the sequential sequence number of the blocks of VGG16 in the encoding layer. Essentially, we found that the deep map generated directly by sampling the deep feature in block 4 is more significant. Specifically, in the CGNet network, CGM 4, CGM 3 and CGM 2 are responsible for feature fusion from deep to shallow respectively. In this process, with the supervision of deep features, CGM can better obtain the receptive field, so as to improve the expression of features. The convolutional block is composed of the convolutional layer, batch normalization, and ReLU activation function. Classifier is a convolutional layer for binary classification.

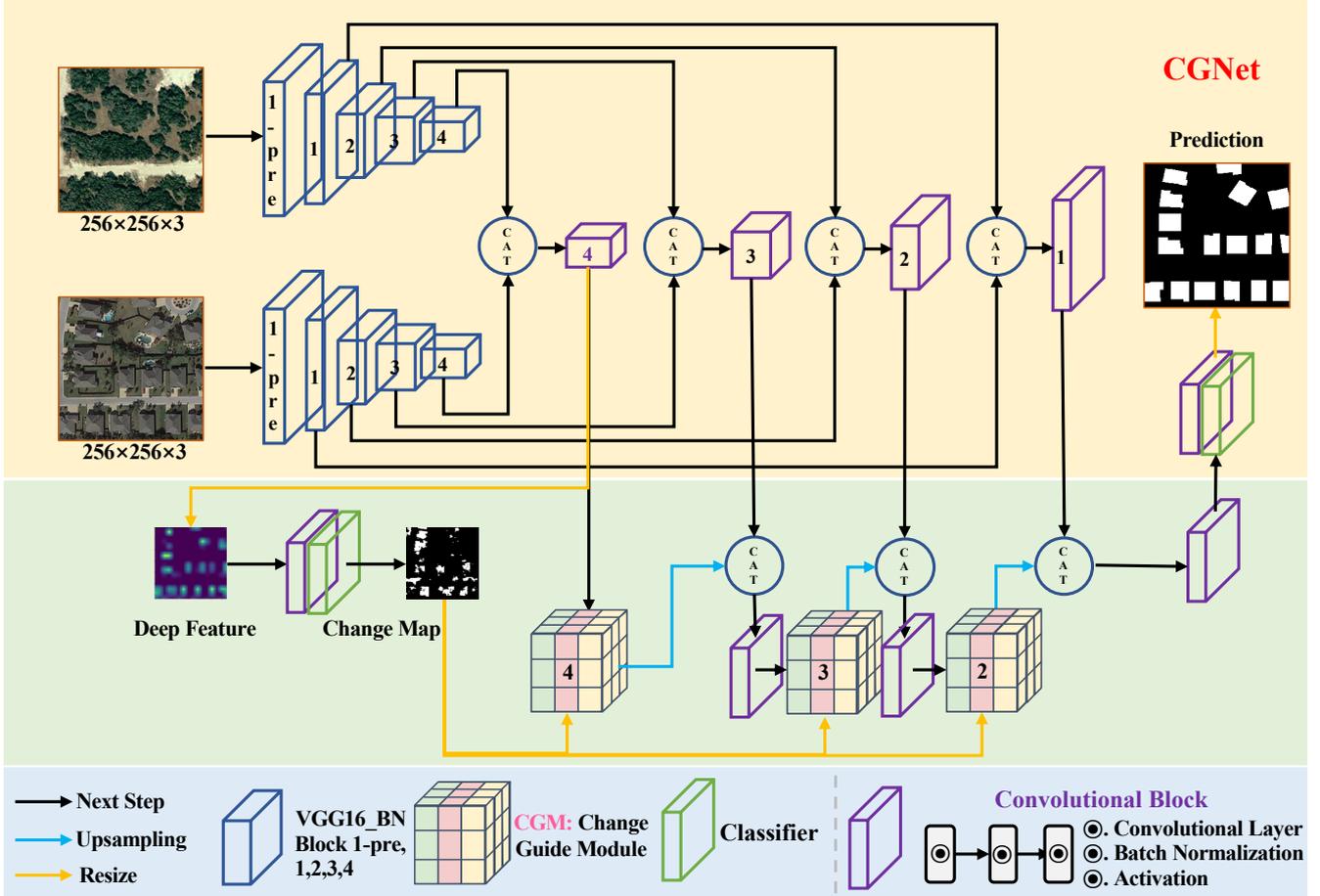

**Fig. 2.** Illustration of the CGNet we proposed.

*B. Change Guide Module*

We generate change maps from deep features with rich semantic information. We broaden the receptive field between pixels by using the proposed CGM module to guide the multi-scale feature fusion process while using the change map as prior knowledge. The objective of the attention mechanism is to ignore the majority of the unimportant information and select a little quantity of important information from a vast amount of information. In the very unbalanced binary classification issue, the attention mechanism for change detection is to locate the significant change features with a small number of pixels and disregard the background information with a large number of pixels. The higher of the weight, the more weight is given to the corresponding value when calculating attentional mechanisms. In other words, the weight represents the importance of the change information and the value represents the corresponding information. Self-attention mechanism is a version of the attention mechanism. It is better at capturing the internal correlations of the data or features and less dependent on external inputs. As shown in Fig. 3, our proposed CGM is a self-attention module, which can obtain significant change information in the size of 256*256 images. Through the feature maps extracted from different stages of our CGNet network, the deep feature maps rich in semantic information generate the changing guide map, which can be used as prior information to guide multi-scale feature fusion, so as to further improve the expression ability of changing features.

The original self-attention [28] is described as:

$$Attention(Q, K, V) = Softmax\left(\frac{QK^T}{\sqrt{d_{head}}}\right)V \quad (1)$$

where $Q$, $K$, and $V$ stand for Query, Key, and Value, respectively. However, it's very computationally intensive because $Q$, $K$, and $V$ all have the dimensions $HW \times C$, where $H$, $W$, and $C$ define the size of the input image. The SoftMax calculation can numerically transform the score of $Q$ and $K$, which can not only normalize the score, but also organize the original calculated score into a probability distribution with the weight of all elements equal to 1, and the heavyweight representing the information with high correlation. Mathematically, given the changing guide map as $f\_GC$. First, we get the weight graph $W_{GC}$ with sigmoid because the sigmoid converts the input to a value of 0-1. The weight graph represents the feature maps extracted at different stages of our CGNet network, which is also the changing guide map in CGM. The size of the value reflects the degree of change. The likelihood of a pixel change increases with increasing weight value. Therefore, the guidance change information can be displayed through the weight graph.

$$W_{GC} = Sigmoid(f\_GC) \quad (2)$$

Given the input feature map as F. And the changing guide map, weighted F_GC can be obtained by multiplying the variable weight graph with the input feature graph. The following is the precise formula:

$$f_{GC} = W_{GC} \cdot conv(\mathcal{F}) \quad (3)$$

$Q$, $K$, and $V$ in our self-attention are mainly calculated in the following ways:

$$(Q, K, V) = (QW_Q, KW_K, VW_V) \quad (4)$$

Where $W_Q$, $W_K$ and $W_V$ are, respectively, the query, key, and value weight matrices of the liner projection. We estimated our attention map as follows:

$$\mathcal{A}_i = softmax\left(\frac{Q_i K_i^T}{\sqrt{d_k}}\right), i = 1, \dots N_h \quad (5)$$

Where $\mathcal{A}_i$ refers to the attention map, $i$ is the $i$-th head in the multi-head self-attention mechanism, which can extract rich potential features, $Q_i$ is the corresponding query matrix calculated in the $i$-th head, and $d_k$ refers to the number of channels in the key matrix. After that, by mutually multiplying the aforementioned $\mathcal{A}_i$ and $V_i$, one may obtain the output feature map of the multi-head self-attention map.

$$\mathcal{H}_i = \mathcal{A}_i V_i \quad (6)$$

Finally, after adding the convolution operation to the Input feature map, we obtain the output of our CGM:

$$output = conv(\mathcal{H}_i) + \mathcal{F} \quad (7)$$

Through the deep feature of Fig.3, it is very significant to introduce the changing guide map into CGM because the changing guide map can help CGM focus on the change information itself.

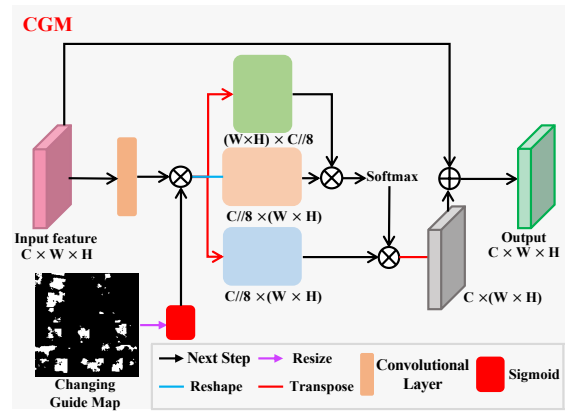

**Fig. 3.** Illustration of our change guide module (CGM).

*C. Model Details*

The CD is an extremely unbalanced binary classification task. Thus, to reduce this imbalance in our experiment, we

use cross-entropy as the loss function. The cross-entropy function is defined as:

$$Loss = -(\hat{y}_i \log(y_i) + (1 - \hat{y}_i) \log(1 - y_i)) \quad (8)$$

where $y_i$ and $\hat{y}_i$ stand in for the label value and the anticipated value of pixel $i$, respectively. The total number of pixels, or $N$, is determined by multiplying the number of pixels in one picture by the batch size. Since the change guiding map can provide change prior information for the decoder part to enhance change feature representation, by computing the cross entropy between the change guiding map with the ground truth, we apply deep supervision to the change guiding map. As a result, the deep supervision loss is added to create the overall loss function.

## III. EXPERIMENT

The experimental datasets, environment, comparison models, and assessment measures are all introduced in this part. Later, a detailed discussion of the experimental findings and the ablation research will be discussed later.

### A. Experimental setup

**LEVIR-CD** [24]: A publicly available large-scale building change detection dataset called LEVIR-CD [24] has 637 very high-resolution (0.5m/pixel) image patch pairings witha 1024 x 1024 pixel size. Significant land-use changes, particularly the expansion of development, can be seen in these bitemporal photos that span a time period of 5 to 14 years. The 20 separate regions that make up the bitemporal pictures on LEVIR-CD [24] are spread across various cities in Texas, including Austin, Lakeway, Bee Cave, Buda, Kyle, Manor, Pflugerville, Dripping Springs, etc. There are 31,066,643 changed pixels and 636,876,269 unchanged pixels are the corresponding numbers. We used the standard data split (445/64/128 image pairs for training, validation, and testing) which was provided on the researchers' web pages. The original image pairs are trimmed to sizes of 256 × 256 without overlapping in order to compare them fairly to other state-of-the-art approaches. Therefore, the final dataset which is depicted in Tab. I, consists of 7120/1024/2048 pairs of patches for training, validation, and testing, respectively.

**WHU-CD** [37]: A huge very high-resolution image pair with a size of over thirty thousand × fifteen thousand pixels that is split into 512 × 512, is included in the open remote sensing buildings change detection dataset known as WHU-CD[37]. It encompasses a region which has very big damage because of the earthquake in February 2011 and was rebuilt in the later years. There are 21,442,501 changed pixels and 481,873,979 unchanged pixels, respectively. We used the standard data split (1260 picture pairings for training; 690 image pairs for testing) that was provided on the researchers' web pages. The original image pairs are trimmed into sizes of 256 × 256 without overlapping in order to compare them fairly to other state-of-the-art approaches. Actually, the validation dataset was created by randomly choosing 10% of the images from the training dataset. As shown in Tab. I, the training set in our dataset includes 4536 pairs, the validation set includes 504 pairs, and the test set includes 2760 pairs.

**SYSU-CD** [38]: A publicly available remote sensing CD dataset called SYSU-CD [38] has 20000 VHR image pairings, which was captured in Hong Kong between 2007 and 2014 and is split into 256 × 256. The SYS-CD dataset contains a number of different sorts of modifications, including (a) newly constructed urban buildings, (b) suburban dilatation, (c) groundwork before construction, (d) change in vegetation, (e) road extension, and (f) sea development. There are 286,092,024 changed pixels and 1,024,627,976 unchanged pixels, respectively. As can be seen in Tab. I, we used the standard data split (the training set has 12000 image pairs, the validation set has 4000 image pairs, and the testing set has 4000 image pairs) that was posted on the researchers' websites.

**S2Looking-CD** [39]: A building-change-detection collection called S2Looking-CD [39] comprises expansive side-looking satellite images taken at different off-nadir angles. It includes about 65,920 annotated examples of angles. It includes about 65,920 annotated examples of changes throughout the world and 5000 bitemporal image pairings of rural regions. By offering (1) wider viewing angles, (2) significant light fluctuations, and (3) the additional complexity of rural photos, it improves upon previous datasets. There are 66,552,990 changed pixels and 5,176,327,010 unchanged pixels, respectively. We used the standard data split (the training set has 3500 image pairings, the validation set has 500 image pairings, and the testing set has 1000 image pairings) that was provided on the researchers' web pages. The original image pairs are trimmed into sizes of 256 × 256 without overlapping in order to compare them fairly to other state-of-the-art approaches. As seen in Tab. I, the dataset for our studies consists of 56,000/8,000/16,000 pairs of patches for training, validation, and testing, respectively.

TABLE I THE STATISTICS OF THE FOUR DATA SETS USED IN THE EXPERIMENT (INCLUDING THE IMBALANCE RATIO)

| Dataset | Spatial Resolution | Size/ Image | Number of Pixels in Datasets | | | Number of Images in Datasets | | |
|---|---|---|---|---|---|---|---|---|
| | | | Changed Pixel | Unchanged Pixel | Imbalanced Ratio | Train | Validation | Test |
| LEVIR | 0.5m/pixel | 256×256 | 31066643 | 636876269 | 1:20.50 | 7120 | 1024 | 2048 |
| WHU | 0.075m/pixel | 256×256 | 21442501 | 481873979 | 1:22.47 | 4536 | 504 | 2760 |
| SYSU | 0.5m/pixel | 256×256 | 286092024 | 1024627976 | 1:3.58 | 12000 | 4000 | 4000 |
| S2Looking | 0.8m/pixel | 256×256 | 66552990 | 5176327010 | 1:77.78 | 56000 | 8000 | 16000 |

**Evaluation Metrics**. We employ the F1-score ($F1$), Precision ($Pre.$), Recall ($Rec.$), and Intersection over Union (IoU) metrics to assess the suggested technique statistically. They may all be used to compare prediction maps and GT, and they are all clearly specified as follows:

$$F1 = \frac{2}{Pre.^{-1} + Rec.^{-1}} \quad (9)$$

$$Recall = \frac{TP}{TP+FN} \quad (10)$$

$$Precison = \frac{TP}{TP+FP} \quad (11)$$

$$IoU = \frac{TP}{TP+FN+FP} \quad (12)$$

where the letters TP, TN, FP, and FN stand for the corresponding totals of true positives, true negatives, false positives, and false negatives. Better CD performance in the experiment is shown by the higher F1 and IoU values.

### B. Experimental environment

We use PyTorch to construct our models, and one NVIDIA RTX 3090 GPU is used to train each model. Using the AdamW [36] optimizer, we lower the loss by setting the weight decay and learning rate to 0.0025 and 0.0005, respectively. To make it easier to compare with other state-of-the-art methods, we crop all the data to 256*256. At the same time, the data was also enhanced using data augmentation techniques including random Gaussian noise, random salt and pepper noise, random cropping, and random rotation of a certain angle. Choosing a batch size and the maximum epoch of eight and fifty for the experiment in order to make the model converge according to several experiments despite the GPU's constraints. In a total of 50 training epochs, we saved the model with the best F1 score and IoU on the validation set for testing.

### C. Comparison with state-of-the-art methods

In order to assess the performance of the proposed CGNet, we contrast a wide range of state-of-the-art CD methods. To the best of our knowledge, we found that the current deep learning models for change detection mainly come in two ways: CNN-based (including attention-based) and transformer-based methods. We selected some state-of-the-art change detection algorithms for high-resolution remote sensing images. Among them, CNN-only methods include FC-EF [20], FC-Siam-conc [20], FC-Siam-diff [20]; Some excellent attention-based methods include IFNet [23], STANet-PAM [24], SNUNet [25], MSPSNet [26], UPerNet [27], HANet [22], HCGMNet [35]; transformer-based methods include BIT [31], Change Former [32] and RSP-BIT [33]. RSP-BIT [33] is a pre-trained model on large-scale remote sensing datasets. The specifical details are as follows:

1) **FC-EF** [20]: An early fusion network that is completely convolutional using the U-Net structure as its foundation, with the concatenation of the two input images from the pair being compared as its input.

2) **FC-Siam-conc** [20]: A model using Siamese concatenation that is completely convolutional. During the decoding step, it concatenates two skip connections, each originating from an encoded stream.

3) **FC-Siam-diff** [20]: A model using a Siamese-difference model that is completely convolutional that, rather than joining the two connections from the encoding streams, concatenates the absolute value of their difference.

4) **IFNet** [23]: Deeply supervised image matching network (IFN) in which highly representative deep features of a two-time image are extracted for the first time through a fully convolutional two-stream architecture and The extracted worm scores are fed into a deeply monitored distinct discriminant network (DDN) for change detection.

5) **STANet-PAM** [24]: A Siamese-based neural network for spatial-temporal attention that completely utilizes the spatial-temporal link to produce features that are both illumination-invariant and misregistration-resistant.

6) **SNUNet** [25]: A densely connected Siamese network based on NestedUNet that gathers and improves characteristics from various semantic levels using

7) **MSPSNet** [26]: A parallel convolutional structure (PCS) based deep multiscale Siamese neural network (SNN) that can integrate various temporal characteristics and further enhance the capability of feature representation through a self-attention model.
8) **UPerNet** [27]: A feature pyramid network-based Unified Perceptual Parsing Network for scene understanding was developed. The FPN is a general feature extractor that takes advantage of multi-level feature representations in a pyramidal hierarchy.
9) **HANet** [22]: The discriminatory symmetrical multi-layer multi-attention network is designed by combining various scales functionality and fine-grained spatial-temporal change function, in which the tiny and fast module of HAN can capture long-term dependencies separate from two dimensions.
10) **HCGMNet** [35]: Change detection using a hierarchical change guiding map network (An initial version of our proposed CGNet).
11) **BIT** [31]: A Bitemporal Image Transformer (BIT) is used to efficiently and effectively describe long-range context inside a bitemporal picture, which aids in recognizing the change of interest and rejecting irrelevant changes.
12) **Change Former** [32]: A lightweight MLP decoder and a hierarchical transformer encoder that can effectively display the multi-scale long-range features needed for accurate CD.
13) **RSP-BIT** [33]: An implementation of the BIT [31] approach using remote sensing pretraining, that uses the transformer-based visual neural backbone network ViTAE [40] as the basic framework and the large data set of MillionAID [41] as pretraining.

It's important to note that we used the aforementioned strategy in an identical environment, with all hyperparameters set to the recommended values on GitHub.

Tab. II and III show the quantization results of our method compared to all other state-of-the-art methods. To facilitate readers to view our experimental results, we use bold red, blue and black to represent the first, second and third place in each column score, respectively. On the four datasets, we can see that our CGNet performs best in terms of F1, OA, and IoU metrics.

For the LEVIR-CD dataset, we can find that our CGNet method achieves the best results on the LEVIR-CD [24] dataset in the key indicators of F1 and IoU, and CGNet achieves first place. Our previously proposed HCGMNet [35], which is the base version of CGNet, can almost achieve second place. This also proves the effectiveness of the proposed change detection framework method based on a hierarchical change guide map.

For the WHU-CD [37] dataset, we find that our proposed change detection framework based on a hierarchical change guide map achieves the first performance in all four evaluation metrics (F1, Pre., Rec., IoU). HCGMNet [35], the basic version of CGNet, has achieved second place, which directly proves the effectiveness of our method again. The overall rule is that the method based on the attention mechanism is generally better than the method based on pure CNN, which is also directly related to the fact that the attention mechanism can obtain a larger receptive field and obtain contextual information. Among them, transformer-based methods (such as Change Former and RSP-BIT) also have certain advantages, but their results are not as good as ours.

The dataset of SYSU-CD [38] not only has a large amount of training data but also are quite challenging. Although the imbalance ratio of the SYSU-CD [38] dataset is only 1:3.58 compared with LEVIR-CD [24] and WHU-CD [37] datasets, its data size is larger and more challenging. By observing that the overall scores of each evaluation index of this dataset are not particularly high, it can also be seen that this dataset is very challenging. Even so, our proposed CGNet still achieves first place in the key F1 and Intersection Over Union (IoU) metrics and second place in Precision (Pre.). This also proves the superiority of our proposed change detection framework based on a hierarchical change guide network (CGNet) on the challenging SYSU-CD [38] dataset.

Especially, S2Looking-CD [39] has an Imbalanced Ratio as high as 77.78, which means the ratio of foreground images (positive samples) to background images (negative samples) is 1:77.78, and further proves it is an extremely unbalanced dataset. This S2Looking-CD [39] dataset is quite challenging by observing that the overall scores of the various evaluation metrics are low. By observation, our proposed CGNet can still achieve the first place in the key F1 and Intersection over Union (IoU) metrics, and the second place in Recall (Rec.). This also proves the superiority of our proposed change detection framework based on a hierarchical change guide map (CGNet) on the extremely challenging S2Looking-CD dataset. In summary, our CGNet still achieves very good results. Numerous studies show how successful our CGNet approach is in extracting change information.

The qualitative visualization outcomes of our technique in comparison to the other state-of-the-art methods are shown in Fig.4-7. Since the data sets of SYSU-CD [38] and S2Looking-CD [39] are relatively large, we randomly show seven visualized test images, while LEVIR-CD [24] and WHU-CD [37] randomly show four visualized test images. In order to be able to give readers a more intuitive view of our experimental results, we denote FP in red and FN in blue. We marked the more challenging change information in all methods in the same position with small red boxes on LEVIR-CD [24] and WHU-CD [37]; We annotated the change information of some methods in certain

TABLE II A COMPARISON OF THE RESULTS WITH OTHER SOTA CHANGE DETECTION METHODS ON LEVIR-CD AND WHU-CD.

| Model | LEVIR-CD [24] | | | | WHU-CD [37] | | | |
|---|---|---|---|---|---|---|---|---|
| | F1 | Pre. | Rec. | IoU | F1 | Pre. | Rec. | IoU |
| FC-EF [20] | 61.52 | 73.31 | 53.00 | 44.43 | 58.05 | 76.49 | 46.77 | 40.89 |
| FC-Siam-conc [20] | 64.41 | **95.30** | 48.65 | 47.51 | 63.99 | 72.06 | 57.55 | 47.05 |
| FC-Siam-diff [20] | 89.00 | 91.76 | 86.40 | 80.18 | 86.31 | 89.63 | 83.22 | 75.91 |
| IFNet [23] | 42.38 | 47.92 | 37.99 | 26.89 | 64.15 | 65.59 | 62.77 | 47.22 |
| STANet-PAM [24] | 85.20 | 80.80 | **90.10** | 74.22 | 82.00 | 75.70 | 89.30 | 69.44 |
| SNUNet [25] | 89.97 | 91.31 | 88.67 | 81.77 | 87.76 | 87.84 | 87.68 | 78.19 |
| MSPSNet [26] | 89.67 | 90.75 | 88.61 | 81.27 | 86.49 | 87.84 | 85.17 | 76.19 |
| UPerNet [27] | 89.72 | 91.65 | 87.87 | 81.36 | 87.71 | 87.89 | 87.52 | 78.10 |
| HANet [22] | **90.28** | 91.21 | 89.36 | **82.27** | **88.16** | 88.30 | 88.01 | **78.82** |
| HCGMNet [35] | **91.77** | 92.96 | **90.61** | **84.79** | **92.08** | 93.93 | **90.31** | **85.33** |
| BIT [31] | 89.94 | 90.33 | 89.56 | 81.72 | 80.97 | 74.01 | 89.37 | 68.02 |
| Change Former [32] | 90.20 | 92.05 | 88.37 | 82.21 | 87.18 | **92.70** | 82.28 | 77.27 |
| RSP-BIT [33] | 89.71 | 92.00 | 87.53 | 81.34 | 78.50 | 69.93 | **89.45** | 64.60 |
| **CGNet (Ours)** | **92.01** | **93.15** | **90.90** | **85.21** | **92.59** | **94.47** | **90.79** | **86.21** |

\* ALL VALUES ARE IN %. TO MAKE IT EASIER TO UNDERSTAND, **FIRST**, **SECOND** AND **THIRD**.

TABLE III A COMPARISON OF THE RESULTS WITH OTHER SOTA CHANGE DETECTION METHODS ON SYSU-CD AND S2LOOKING-CD.

| Model | SYSU-CD [38] | | | | S2Looking-CD [39] | | | |
|---|---|---|---|---|---|---|---|---|
| | F1 | Pre. | Rec. | IoU | F1 | Pre. | Rec. | IoU |
| FC-EF [20] | 68.20 | 75.27 | 62.35 | 51.75 | 7.65 | **81.36** | 8.95 | - |
| FC-Siam-conc [20] | 73.38 | 81.57 | 66.69 | 57.96 | 13.54 | 68.27 | 18.52 | - |
| FC-Siam-diff [20] | 69.11 | **91.27** | 55.61 | 52.80 | 46.60 | 83.49 | 32.32 | 30.38 |
| IFNet [23] | 63.57 | 50.56 | **85.61** | 46.60 | 26.07 | 21.39 | 33.35 | 14.99 |
| STANet-PAM [24] | 77.80 | 77.60 | **78.10** | 63.71 | 47.50 | 36.40 | **68.20** | 31.13 |
| SNUNet [25] | **78.50** | 76.91 | **80.15** | 64.60 | 47.78 | 45.25 | 50.60 | 31.39 |
| MSPSNet [26] | 76.79 | 77.29 | 76.29 | 62.32 | 55.20 | 58.05 | 52.61 | 38.12 |
| UPerNet [27] | 77.61 | 81.13 | 74.38 | 63.41 | 58.05 | 70.18 | 49.50 | 40.90 |
| HANet [22] | 77.41 | 78.71 | 76.14 | 63.14 | 58.54 | 61.38 | 55.94 | 41.38 |
| HCGMNet [35] | **79.42** | 84.31 | 75.07 | **65.86** | **62.96** | 70.51 | **56.87** | **45.94** |
| BIT [31] | 73.32 | 75.15 | 71.58 | 57.88 | 62.65 | 70.26 | 56.53 | **45.62** |
| Change Former [32] | 64.62 | 59.40 | 70.84 | 47.73 | **63.39** | 72.82 | 56.13 | - |
| RSP-BIT [33] | 78.26 | **86.29** | 71.61 | 64.29 | 55.52 | 58.06 | 53.19 | 38.42 |
| **CGNet (Ours)** | **79.92** | **86.37** | 74.37 | **66.55** | **64.33** | 70.18 | **59.38** | **47.41** |

\* ALL VALUES ARE IN %. TO MAKE IT EASIER TO UNDERSTAND, **FIRST**, **SECOND** AND **THIRD**.

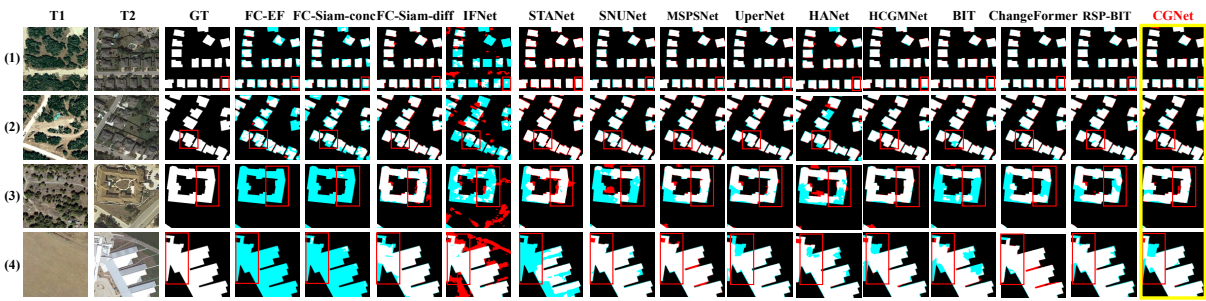

**Fig. 4.** Qualitative experimental results on LEVIR-CD [24]. TP (white), TN (black), FP (red), and FN (blue).

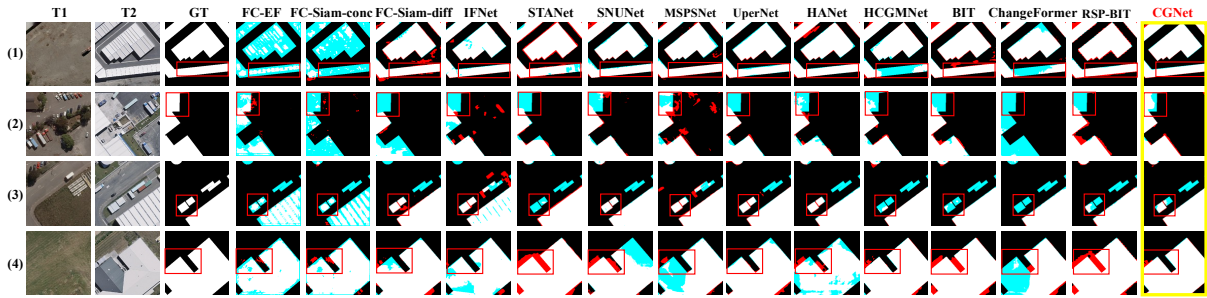

**Fig. 5.** Qualitative experimental results on WHU-CD [37].

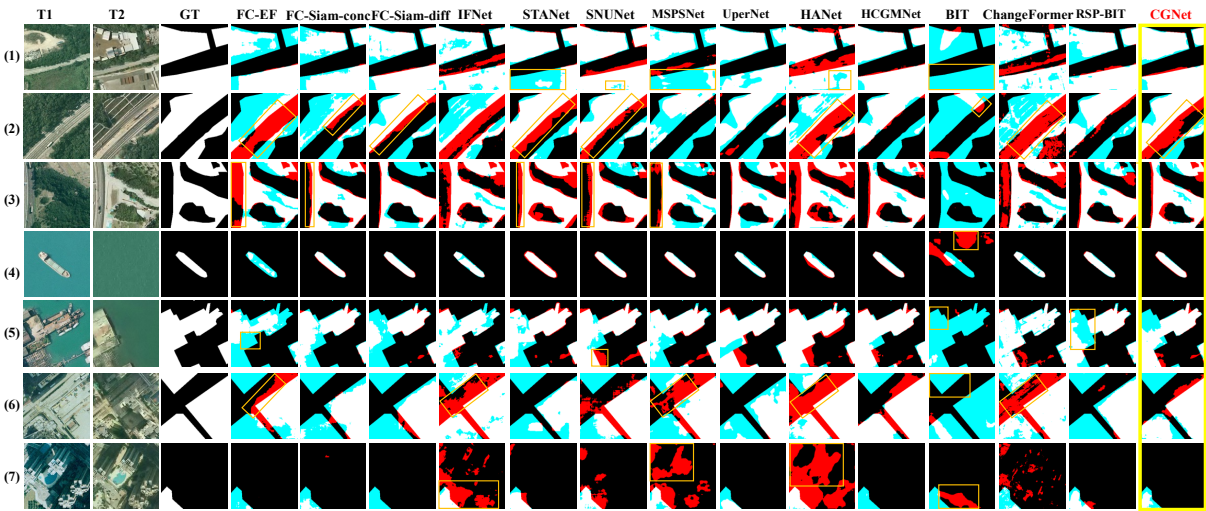

**Fig. 6.** Qualitative experimental results on SYSU-CD [38].

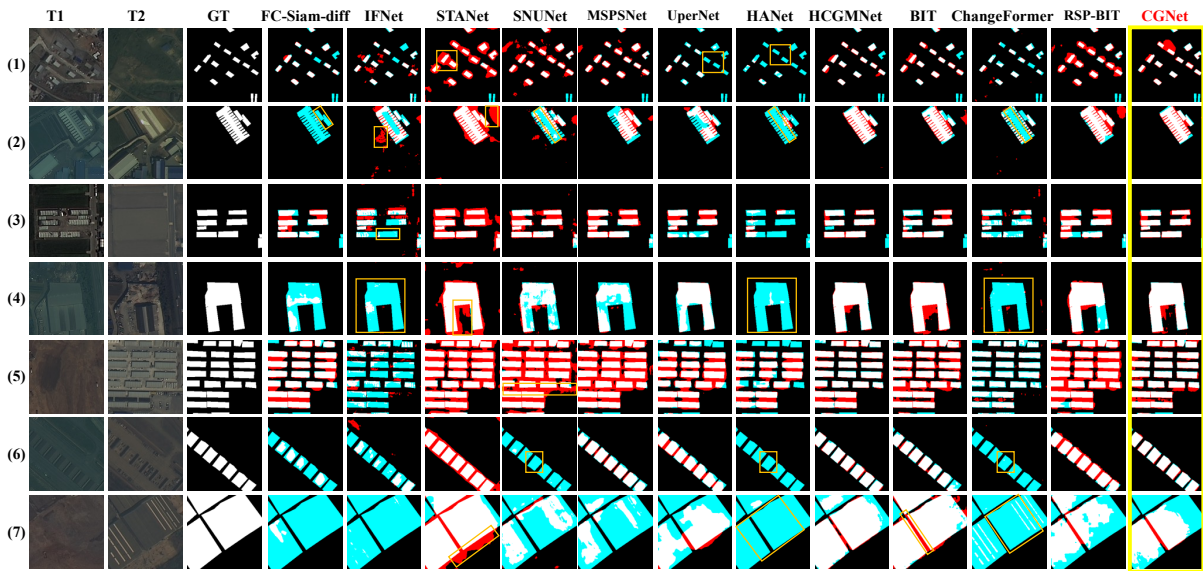

**Fig.7.** Qualitative experimental results on S2Looking-CD [39].

challenging positions with small yellow boxes on SYSU-CD [38] and S2Looking-CD [39]. We can find that the feature differences between the four data sets are still quite significant. In the process of feature learning, the model needs to have a strong feature-learning ability to achieve good results.

For the LEVIR-CD [24] dataset, the classical change detection methods FC-EF and FC-Siam-conc have some missed detections in Fig. 4 (3,4), IFNet has some false detections, and some methods with attention mechanisms still cannot solve the phenomenon of internal holes in change information in the detection results, such as the irregular buildings in Fig. 4 (3).

For the WHU-CD [37] dataset, it is worth noting that the spatial resolution of the WHU-CD dataset (0.075m/ pixel) is very high, which means that the background of key features in the whole scene is more complex, the interior texture of the building is clearer, the edge details are richer, the vehicles with the same colour compared to the overall colour of the building, and the ground is also more likely to be mixed up. This directly increases the difficulty of change detection. Through observation, we find that the results of qualitative analysis and quantitative analysis are consistent. If the F1 and IoU values of quantitative results are low methods, then the results of qualitative visualization areprone to false detection (red area) and missed detection (blue area) phenomena. Even so, our proposed CGNet can perform well.

For the SYSU-CD[38] dataset, we find that there are more red and blue areas in the whole SYSU-CD data set, which means that the phenomenon of false detection and missed detection is serious, and the edge of change information is incomplete and there are a lot of holes inside, which indicates that this data set is relatively challenging. We analyze the reason for this because the SYSU-CD [38] dataset has many main types of changes. Among them, the detection effect of Fig.6 (4) is relatively good, because the background of the sea is relatively simple, which is not only because of seasonal changes, light intensity and other factors, the sea colour in the same area is not the same. In addition to this type of data, other scene backgrounds are relatively complex, and the feature similarity of the change region is high and the shape is irregular, so the detection difficulty is naturally high. In general, our proposed CGNet has the best visualization effect and can solve the phenomenon of incomplete edge information and internal holes in the change region to a certain extent.

For the S2Lookeing-CD [39] dataset, we find that there are more red and blue regions on the whole S2Looking-CD dataset, which are extremely challenging datasets like the SYSU-CD [38] dataset. By analyzing the data of T1 and T2, it is not difficult to find that there is a big difference in colour and an obvious difference in texture features between them. For the change regions with small targets as shown in Fig.7 (1), with rich edge information as shown in Fig.7 (2), and with relatively dense arrangements as shown in Fig.7 (3,5,6) are still difficult problems in the change detection task. In general, our proposed change detection framework based on a hierarchical change guidance map (CGNet) has fewer false detections (red areas) and missed detections (blue areas) on the S2Looking-CD [39] dataset, indicating that our method can solve the phenomenon of incomplete edge information and internal holes in the change region to a certain extent. The numerous visualization results show that our proposed CGNet can solve the phenomenon of incomplete edge information and internal holes in the change region.

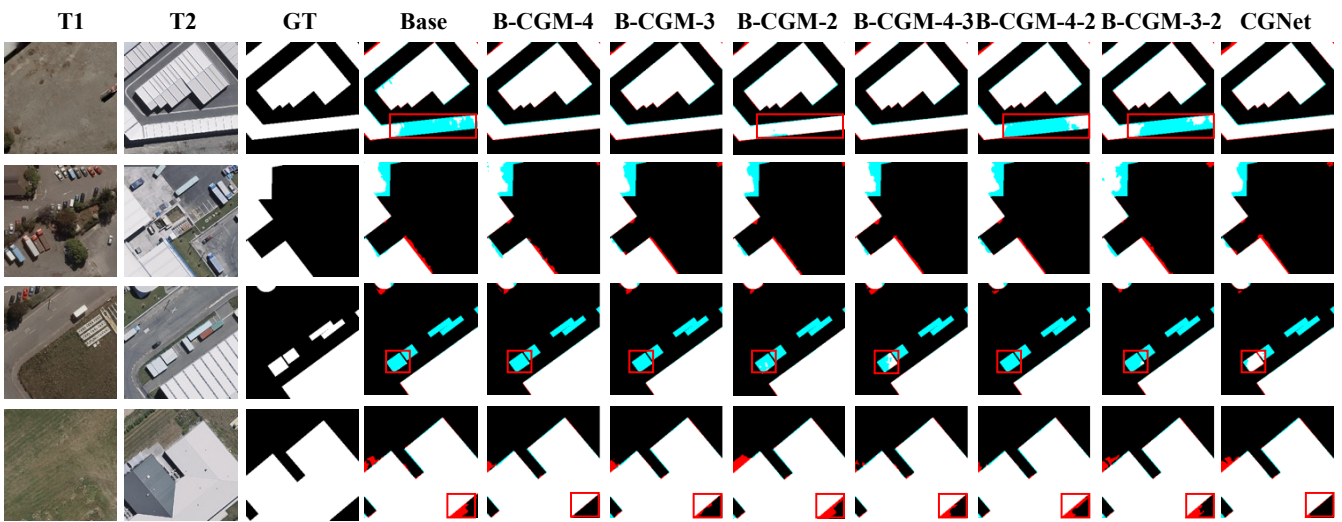

**Fig. 8.** Qualitative results of ablation study on WHU-CD dataset.

## C. Ablation Study

To verify the efficacy of our suggested model, we set up the following models. In this ablation experiment, CGM-4, CGM-3 and CGM-2 respectively represent the CGMs at different stages in Fig.2.

- **Base model**: The VGG 16_BN serves as the baseline model (No CGM).
- **CGNet**: Base model + CGM 4 + CGM 3 + CGM 2.
- **B-CGM-4**: Base model + CGM 4.
- **B-CGM-3**: Base model + CGM 3.
- **B-CGM-2**: Base model + CGM 2.
- **B-CGM-4-3**: Base model + CGM 4 + CGM 3.
- **B-CGM-4-2**: Base model + CGM 4 + CGM 2.
- **B-CGM-3-2**: Base model + CGM 3 + CGM 2.

**Ablation on single CGM.** We verify the effectiveness of CGM by comparing the effect of a single CGM at different stages in the proposed CGNet network. The three stages of CGM are called CGM 4, CGM 3, and CGM 2 respectively. It can be observed from the Tab. IV that in any data set, the base model without CGM has the lowest accuracy, the CGNet model with three CGMs has the highest accuracy, and the accuracy of one CGM ranks in the middle. As shown in Fig. 8, the results of the visualization are consistent with those of the quantification. These large number of experiments fully prove that our proposed CGM module has better characteristics of extracting change information.

TABLE IV

ABLATION STUDY ON THE MAIN CGNET MODULE COMPONENT (SINGLE CGM AND DOUBLE CGM) ON FOUR DATASETS. F1-SCORE IS COMPARED.

| Model | F1 | | | |
|---|---|---|---|---|
| | LEVIR | WHU | SYSU | S2Looking |
| Base | 91.61 | 91.78 | 79.45 | 63.07 |
| B-CGM-4 | 91.89 | 92.28 | 79.57 | 62.79 |
| B-CGM-3 | 91.86 | 92.20 | 79.53 | 63.43 |
| B-CGM-2 | 91.77 | 91.87 | 79.72 | 62.86 |
| B-CGM-4-3 | 91.88 | 92.04 | 78.82 | 63.95 |
| B-CGM-4-2 | 91.88 | 92.15 | 79.81 | 63.67 |
| B-CGM-3-2 | 91.76 | 91.36 | 79.85 | 63.47 |
| **CGNet** | **92.01** | **92.59** | **79.92** | **64.33** |

**Ablation on double CGMs.** We verify the effectiveness of CGM by comparing the effects of double CGMs at different stages in the proposed CGNet network. It can be observed from the Tab. IV that the base model without CGM has the lowest accuracy in any data set, the CGNet model with three CGMS has the highest accuracy, and the accuracy with double CGM ranks in the middle. At the same time, as shown in Tab. IV we can find that the effect with double CGMs is better than that with single CGMs. In general, our proposed model, CGNet network, also achieved gradually increased effects when there were 1, 2, and 3 CGMs respectively in the network. We can find that the results of the visualization are consistent with those of the quantification from Fig.8. Obviously, the basic model has an obvious change in information edge incomplete and internal hole phenomenon without the CGM module. With the addition of a single CGM module in different stages of the network, the extraction effect of change information has been better improved. When two CGM modules are added to the network at different stages, the internal hole phenomenon is significantly improved as shown in Fig.8. When three CGM modules are added to the network, our proposed change detection framework based on hierarchical change guidance graph (CGNet) can solve the problem of incomplete edge and internal hole phenomenon of change information very well. Therefore, these large numbers of experiments also fully prove that our proposed CGM module has better characteristics for extracting change information.

TABLE V

ABLATION STUDY ON THE TRAINING AND INFERENCING TIME ON WHU-CD DATASET. S MEANS SECONDS.

| Model | Training time (1 Epoch / S) | Inference time (Test set / S) |
|---|---|---|
| FC-EF [20] | 31.80 | 77.00 |
| FC-Siam-conc [20] | 36.21 | 101.61 |
| FC-Siam-diff [20] | 36.73 | 76.86 |
| IFNet [23] | 117.24 | 518.62 |
| STANet-PAM [24] | 195.24 | 230.38 |
| SNUNet [25] | 402.37 | 123.27 |
| MSPSNet [26] | 802.38 | 131.16 |
| UPerNet [27] | 39.39 | 90.42 |
| HANet [22] | 983.86 | 129.02 |
| HCGMNet [35] | 324.95 | 54.34 |
| BIT [31] | 145.24 | 23.93 |
| Change Former [32] | 370.43 | 71.23 |
| RSP-BIT [33] | 205.01 | 116.65 |
| **CGNet (Ours)** | **135.76** | **33.15** |

**Ablation on training and inference time.** We compare the training speed of one epoch on the WHU-CD [37] dataset and the inferencing time on the test set. It can be seen from the Tab. IV that from the perspective of training time, some pure CNN structure models such as FC-EF [20], FC-Siam-conc [20], FC-Siam-diff [20] run faster, but the accuracy of the model is low. Some methods that add attention mechanisms such as STANet [24], SNUNet [25], HANet [22], etc., improve the accuracy of the model to a certain extent compared with the pure CNN method, but the consequence of the model becoming more complex is that the running time also increases a lot. Transformer-based methods such as BIT [31] are relatively fast because they have fewer parameters; Change Former [32] is a pure transformer architecture, which is also a bit slower; In general, our proposed CGNet model can ensure both the accuracy of the model and the training efficiency of the model. It has good competitiveness in the two dimensions of time and accuracy. In terms of inference time, our CGNet still has a good performance, and this advantage is more obvious in SYSU-CD [38] and S2Looking -CD [39] with a relatively large amount of data.

## IV. CONCLUSION

In this research, we come to the conclusion that the two primary categories of existing deep learning models for change detection are: CNN-based (including attention mechanism) and transformer-based methods. However, many of them also have insufficient expression problems of change features in the conventional U-Net structure adopted in these methods, which causes inaccurate edge detection and internal holes. To tackle these problems, we improve the conventional U-net architecture and propose a hierarchical change guiding network (CGNet), which uses a deep change guiding feature as prior information to guide multi-scale feature fusion. And we propose a CGM module to get a big receptive field which is a self-attention module. We compare the 13 state-of-the-art CD methods on 4 datasets, and many quantitative and visual findings demonstrate the advantages of our proposed CGNet. Additionally, it is proved that the proposed CGM module has a good impact through many ablation studies on four datasets. In the future, this kind of deep change information to guide the model to fuse multi-scale information is still worth exploring in other state-of-the-art CD methods.

## REFERENCES

[1] A. Singh, "Review Article Digital change detection techniques using remotely-sensed data," *Int. J. Remote Sens.*, vol. 10, no. 6, pp. 989–1003, Jun. 1989, doi: 10.1080/01431168908903939.

[2] G. Xian and C. Homer, "Updating the 2001 National Land Cover Database Impervious Surface Products to 2006 using Landsat Imagery Change Detection Methods," *Remote Sens. Environ.*, vol. 114, no. 8, pp. 1676–1686, Aug. 2010, doi: 10.1016/j.rse.2010.02.018.

[3] P. K. Mishra, A. Rai, and S. C. Rai, "Land use and land cover change detection using geospatial techniques in the Sikkim Himalaya, India," *Egypt. J. Remote Sens. Space Sci.*, vol. 23, no. 2, pp. 133–143, Aug. 2020, doi: 10.1016/j.ejrs.2019.02.001.

[4] C. Song, B. Huang, L. Ke, and K. S. Richards, "Remote sensing of alpine lake water environment changes on the Tibetan Plateau and surroundings: A review," *ISPRS J. Photogramm. Remote Sens.*, vol. 92, pp. 26–37, Jun. 2014, doi: 10.1016/j.isprsjprs.2014.03.001.

[5] N. A. QUARMBY and J. L. CUSHNIE, "Monitoring urban land cover changes at the urban fringe from SPOT HRV imagery in south-east England," *Int. J. Remote Sens.*, vol. 10, no. 6, pp. 953–963, Jun. 1989, doi: 10.1080/01431168908903937.

[6] P. J. HOWARTH and G. M. WICKWARE, "Procedures for change detection using Landsat digital data," *Int. J. Remote Sens.*, vol. 2, no. 3, pp. 277–291, Jul. 1981, doi: 10.1080/01431168108948362.

[7] O. A. Carvalho Júnior, R. F. Guimarães, A. R. Gillespie, N. C. Silva, and R. A. T. Gomes, "A New Approach to Change Vector Analysis Using Distance and Similarity Measures," *Remote Sens.*, vol. 3, no. 11, pp. 2473–2493, Nov. 2011, doi: 10.3390/rs3112473.

[8] A. A. Nielsen, K. Conradsen, and J. J. Simpson, "Multivariate Alteration Detection (MAD) and MAF Postprocessing in Multispectral, Bitemporal Image Data: New Approaches to Change Detection Studies," *Remote Sens. Environ.*, vol. 64, no. 1, pp. 1–19, Apr. 1998, doi: 10.1016/S0034-4257(97)00162-4.

[9] J. S. Deng, K. Wang, Y. H. Deng, and G. J. Qi, "PCA-based land-use change detection and analysis using multitemporal and multisensor satellite data," *Int. J. Remote Sens.*, vol. 29, no. 16, pp. 4823–4838, Aug. 2008, doi: 10.1080/01431160801950162.

[10] C. Wu, B. Du, and L. Zhang, "Slow Feature Analysis for Change Detection in Multispectral Imagery," *IEEE Trans. Geosci. Remote Sens.*, vol. 52, no. 5, pp. 2858–2874, May 2014, doi: 10.1109/TGRS.2013.2266673.

[11] J. Im and J. Jensen, "A change detection model based on neighborhood correlation image analysis and decision tree classification," *Remote Sens. Environ.*, vol. 99, no. 3, pp. 326–340, Nov. 2005, doi: 10.1016/j.rse.2005.09.008.

[12] J. Yang, B. Du, Y. Xu, and L. Zhang, "Can Spectral Information Work While Extracting Spatial Distribution? — An Online Spectral

Information Compensation Network for HSI Classification," *IEEE Trans. Image Process.*, pp. 1–1, 2023, doi: 10.1109/TIP.2023.3244414.

[13] H. Guo, Q. Shi, A. Marinoni, B. Du, and L. Zhang, "Deep building footprint update network: A semi-supervised method for updating existing building footprint from bi-temporal remote sensing images," *Remote Sens. Environ.*, vol. 264, p. 112589, Oct. 2021, doi: 10.1016/j.rse.2021.112589.

[14] H. Guo, B. Du, L. Zhang, and X. Su, "A coarse-to-fine boundary refinement network for building footprint extraction from remote sensing imagery," *ISPRS J. Photogramm. Remote Sens.*, vol. 183, pp. 240–252, Jan. 2022, doi: 10.1016/j.isprsjprs.2021.11.005.

[15] Y. Xu, L. Zhang, B. Du, and L. Zhang, "Hyperspectral Anomaly Detection Based on Machine Learning: An Overview," *IEEE J. Sel. Top. Appl. Earth Obs. Remote Sens.*, vol. 15, pp. 3351–3364, 2022, doi: 10.1109/JSTARS.2022.3167830.

[16] M. Hu, C. Wu, B. Du, and L. Zhang, "Binary Change Guided Hyperspectral Multiclass Change Detection," *IEEE Trans. Image Process.*, vol. 32, pp. 791–806, 2023, doi: 10.1109/TIP.2022.3233187.

[17] J. Li, W. He, and H. Zhang, "Towards Complex Backgrounds: A Unified Difference-Aware Decoder for Binary Segmentation." arXiv, Oct. 26, 2022. Accessed: Apr. 10, 2023. [Online]. Available: http://arxiv.org/abs/2210.15156

[18] H. Chen, N. Yokoya, C. Wu, and B. Du, "Unsupervised Multimodal Change Detection Based on Structural Relationship Graph Representation Learning." Oct. 03, 2022. doi: 10.1109/TGRS.2022.3229027.

[19] C. Wu, B. Du, and L. Zhang, "Fully Convolutional Change Detection Framework with Generative Adversarial Network for Unsupervised, Weakly Supervised and Regional Supervised Change Detection," *IEEE Trans. Pattern Anal. Mach. Intell.*, pp. 1–15, 2023, doi: 10.1109/TPAMI.2023.3237896.

[20] R. C. Daudt, B. L. Saux, and A. Boulch, "Fully Convolutional Siamese Networks for Change Detection." arXiv, Oct. 19, 2018. Accessed: Oct. 04, 2022. [Online]. Available: http://arxiv.org/abs/1810.08462

[21] L. Mou, L. Bruzzone, and X. X. Zhu, "Learning Spectral-Spatial-Temporal Features via a Recurrent Convolutional Neural Network for Change Detection in Multispectral Imagery." Mar. 07, 2018. doi: 10.1109/TGRS.2018.2863224.

[22] C. Han, C. Wu, H. Guo, M. Hu, and H. Chen, "HANet: A hierarchical attention network for change detection with bi-temporal very-high-resolution remote sensing images," *IEEE J. Sel. Top. Appl. Earth Obs. Remote Sens.*, pp. 1–17, 2023, doi: 10.1109/JSTARS.2023.3264802.

[23] C. Zhang *et al.*, "A deeply supervised image fusion network for change detection in high resolution bi-temporal remote sensing images," *ISPRS J. Photogramm. Remote Sens.*, vol. 166, pp. 183–200, Aug. 2020, doi: 10.1016/j.isprsjprs.2020.06.003.

[24] H. Chen and Z. Shi, "A Spatial-Temporal Attention-Based Method and a New Dataset for Remote Sensing Image Change Detection," *Remote Sens.*, vol. 12, no. 10, p. 1662, May 2020, doi: 10.3390/rs12101662.

[25] S. Fang, K. Li, J. Shao, and Z. Li, "SNUNet-CD: A Densely Connected Siamese Network for Change Detection of VHR Images," *IEEE Geosci. Remote Sens. Lett.*, vol. 19, pp. 1–5, 2022, doi: 10.1109/LGRS.2021.3056416.

[26] Q. Guo, J. Zhang, S. Zhu, C. Zhong, and Y. Zhang, "Deep Multiscale Siamese Network With Parallel Convolutional Structure and Self-Attention for Change Detection," *IEEE Trans. Geosci. Remote Sens.*, vol. 60, pp. 1–12, 2022, doi: 10.1109/TGRS.2021.3131993.

[27] T. Xiao, Y. Liu, Y. Jiang, B. Zhou, and J. Sun, "Unified Perceptual Parsing for Scene Understanding".

[28] A. Vaswani *et al.*, "Attention is All you Need," p. 11.

[29] A. Dosovitskiy *et al.*, "AN IMAGE IS WORTH 16X16 WORDS: TRANSFORMERS FOR IMAGE RECOGNITION AT SCALE," 2021.

[30] C. Zhang, L. Wang, S. Cheng, and Y. Li, "SwinSUNet: Pure Transformer Network for Remote Sensing Image Change Detection," *IEEE Trans. Geosci. Remote Sens.*, vol. 60, pp. 1–13, 2022, doi: 10.1109/TGRS.2022.3160007.

[31] H. Chen, Z. Qi, and Z. Shi, "Remote Sensing Image Change Detection With Transformers," *IEEE Trans. Geosci. Remote Sens.*, vol. 60, pp. 1–14, 2022, doi: 10.1109/TGRS.2021.3095166.

[32] W. G. C. Bandara and V. M. Patel, "A Transformer-Based Siamese Network for Change Detection." arXiv, Sep. 01, 2022. Accessed: Dec. 10, 2022. [Online]. Available: http://arxiv.org/abs/2201.01293

[33] D. Wang, J. Zhang, B. Du, G.-S. Xia, and D. Tao, "An Empirical Study of Remote Sensing Pretraining." arXiv, May 15, 2022. Accessed: Dec. 10, 2022. [Online]. Available: http://arxiv.org/abs/2204.02825

[34] Q. Li, R. Zhong, X. Du, and Y. Du, "TransUNetCD: A Hybrid Transformer Network for Change Detection in Optical Remote-Sensing Images," *IEEE Trans. Geosci. Remote Sens.*, vol. 60, pp. 1–19, 2022, doi: 10.1109/TGRS.2022.3169479.

[35] C. Han, C. Wu, and B. Du, "HCGMNET: A HIERARCHICAL CHANGE GUIDING MAP NETWORK FOR CHANGE DETECTION".

[36] I. Loshchilov and F. Hutter, "Decoupled Weight Decay Regularization." arXiv, Jan. 04, 2019. Accessed: Apr. 14, 2023. [Online]. Available: http://arxiv.org/abs/1711.05101

[37] S. Ji, S. Wei, and M. Lu, "Fully Convolutional Networks for

Multisource Building Extraction From an Open Aerial and Satellite Imagery Data Set," *IEEE Trans. Geosci. Remote Sens.*, vol. 57, no. 1, pp. 574–586, Jan. 2019, doi: 10.1109/TGRS.2018.2858817.

[38] Q. Shi, M. Liu, S. Li, X. Liu, F. Wang, and L. Zhang, "A Deeply Supervised Attention Metric-Based Network and an Open Aerial Image Dataset for Remote Sensing Change Detection," *IEEE Trans. Geosci. Remote Sens.*, vol. 60, pp. 1–16, 2022, doi: 10.1109/TGRS.2021.3085870.

[39] L. Shen *et al.*, "S2Looking: A Satellite Side-Looking Dataset for Building Change Detection," *Remote Sens.*, vol. 13, no. 24, p. 5094, Dec. 2021, doi: 10.3390/rs13245094.

[40] Y. Xu, Q. Zhang, J. Zhang, and D. Tao, "ViTAE: Vision Transformer Advanced by Exploring Intrinsic Inductive Bias." arXiv, Dec. 23, 2021. Accessed: Dec. 11, 2022. [Online]. Available: http://arxiv.org/abs/2106.03348

[41] Y. Long *et al.*, "On Creating Benchmark Dataset for Aerial Image Interpretation: Reviews, Guidances and Million-AID." arXiv, Mar. 30, 2021. Accessed: Dec. 11, 2022. [Online]. Available: http://arxiv.org/abs/2006.12485

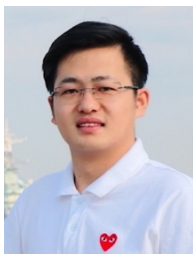

**Chengxi Han** (Student Member, IEEE) received the B.S. degree in remote sensing science and technology from the School of Geosciences and Info-Physics, Central South University, Changsha, China, in 2018. He is currently working toward the Ph.D. degree in photogrammetry and remote sensing with the State Key Laboratory of Information Engineering in Surveying, Mapping, and Remote Sensing, Wuhan University, Wuhan, China.

He was a Trainee with the United Nations Satellite Centre of the United Nations Institute for Training and Research. His research interests include deep learning and remote sensing image change detection. Mr. Han has been the IEEE GRSS Wuhan Student Branch Chapter Chair since 2021. He was the recipient of the IEEE GRSS 2022 Student Chapter Excellence Award. His personal website is https://chengxihan.github.io/.

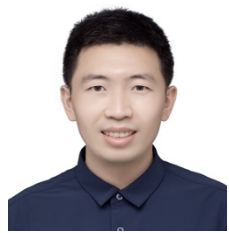

**Chen Wu** (M'16) received B.S. degree in surveying and mapping engineering from Southeast University, Nanjing, China, in 2010, and received the Ph.D. degree in Photogrammetry and Remote Sensing from State Key Lab of Information Engineering in Surveying, Mapping and Remote sensing, Wuhan University, Wuhan, China, in 2015.

He is currently a Professor with the State Key Laboratory of Information Engineering in Surveying, Mapping and Remote Sensing, Wuhan University, Wuhan, China. His research interests include multitemporal remote sensing image change detection and analysis in multispectral and hyperspectral images.

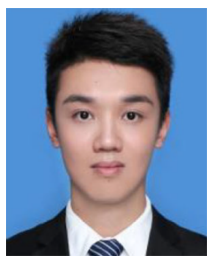

**Haonan Guo** received the B.S. degree in Sun Yat-sen University, Guangzhou, China, in 2020. He is currently working toward the Ph.D. degree with the State Key Laboratory of Information Engineering in Surveying, Mapping, and Remote Sensing, Wuhan University, Wuhan, China.

His research interests include deep learning, building footprint extraction, urban remote sensing, and multisensor image processing.

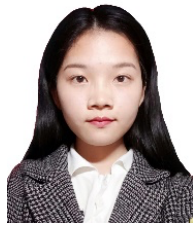

**Meiqi Hu** (Graduate Student Member, IEEE) received the B.S. degree in surveying and mapping engineering from the School of Geoscience and info-physics, Central South University, Changsha, China, in 2019. She is currently pursuing the Ph.D. degree with the State Key Laboratory of Information Engineering in Surveying, Mapping, and Remote sensing, Wuhan University, Wuhan, China.

Her research interests include deep learning and multitemporal remote sensing image change detection.

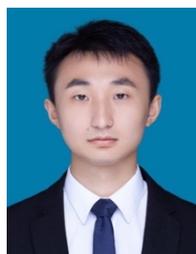

**Jiepan Li** received B.S. degree from School of Electronic Information, Wuhan University, Wuhan, China, in 2016. He currently pursuing the Ph.D. degree at State Key Laboratory of Information Engineering in Surveying, Mapping and Remote Sensing (LIESMARS), Wuhan University.

His research interests include deep learning, building extraction, change detection, salient object detection, and camouflaged object detection.

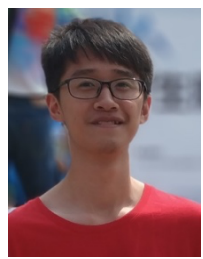

**Hongruixuan Chen** (Student Member, IEEE) received the B.E. degree in surveying and mapping engineering from the School of Resources and Environmental Engineering, Anhui University, Hefei, China, in 2019, and the M.E. degree in photogrammetry and remote sensing from State Key Laboratory of Information Engineering in Surveying, Mapping and Remote Sensing, Wuhan University, Wuhan, China, in 2022. He is now pursuing his Ph.D. degree at the Graduate School of Frontier Science, The University of Tokyo, Chiba, Japan. His current research fields include deep learning, domain adaptation, and multimodal remote sensing image interpretation and analysis. He was a Trainee at the United Nations Satellite Centre (UNOSAT) of the United Nations Institute for Training and Research (UNITAR). He also acts as a reviewer for eight international journals, e.g. IEEE TIP, TGRS, GRSL, and JSTARS.